\documentclass[letterpaper, 10 pt, conference]{ieeeconf}  % Comment this line out if you need a4paper

\IEEEoverridecommandlockouts                              % This command is only needed if
\usepackage{cite}
\usepackage{amsmath}
\usepackage{multirow}

\usepackage{bm}
\usepackage{pifont}
\usepackage{verbatim}
\usepackage{bigstrut,multirow,rotating}
\usepackage{booktabs}
\usepackage{multicol}
\usepackage{color}
\usepackage{mathrsfs}
\usepackage{textcomp}
\usepackage[]{algorithm2e}
\usepackage{mathrsfs,soul}
\usepackage{graphicx}
\usepackage{makecell}
\usepackage{algpseudocode}

\makeatletter

\newcommand{\Rmnum}[1]{\expandafter\@slowromancap\romannumeral #1@}
\makeatother

\IEEEpeerreviewmaketitle
\graphicspath{{Figure/}}
\nocite{*}

\title{\LARGE \bf
Incremental Few-Shot Object Detection for Robotics
}

\author{
	Yiting Li, %\IEEEmembership{Member,~IEEE,}
	Haiyue Zhu$^{\ast}$, %\IEEEmembership{Student Member,~IEEE,}
	Sichao Tian,
	Fan Feng,
	Jun Ma,
	Chek Sing Teo,\\ %\IEEEmembership{Student Member,~IEEE,}
	Cheng Xiang,
	Prahlad Vadakkepat,
	and Tong Heng Lee
	\thanks{Y. Li, S. Tian, F. Feng, C. Xiang, P. Vadakkepat, and T. H. Lee are with National University of Singapore, Singapore 117583 (e-mail: yiting\_li@u.nus.edu, tiansichao@u.nus.edu, e0554256@u.nus.edu, elexc@nus.edu.sg, prahlad@nus.edu.sg, eleleeth@nus.edu.sg).}
	\thanks{H. Zhu and C. S. Teo are with the Adaptive Robotics and Mechatronics Group, Singapore Institute of Manufacturing Technology, Agency for Science, Technology and Research, Singapore 138634 (zhu\_haiyue@simtech.a-star.edu.sg, csteo@simtech.a-star.edu.sg).}
	\thanks{J. Ma is with the Robotics and Autonomous Systems Thrust, The Hong Kong University of Science and Technology (Guangzhou), Guangzhou, China, and the Department of Electronic and Computer Engineering, The Hong Kong University of Science and Technology, Hong Kong SAR, China (e-mail: jun.ma@ust.hk).}
\thanks{$^{\ast}$Corresponding Author: Haiyue Zhu}
}

\begin{document}

\maketitle
\thispagestyle{empty}
\pagestyle{empty}

%%%%%%%%%%%%%%%%%%%%%%%%%%%%%%%%%%%%%%%%%%%%%%%%%%%%%%%%%%%%%%%%%%%%%%%%%%%%%%%%
\begin{abstract}
Incremental few-shot learning is highly expected for practical robotics applications. On one hand, robot is desired to learn new tasks quickly and flexibly using only few annotated training samples; on the other hand, such new additional tasks should be learned in a continuous and incremental manner without forgetting the previous learned knowledge dramatically. In this work, we propose a novel Class-Incremental Few-Shot Object Detection (CI-FSOD) framework that enables deep object detection network to perform effective continual learning from just few-shot samples without re-accessing the previous training data. We achieve this by equipping the widely-used Faster-RCNN detector with three elegant components. Firstly, to best preserve performance on the pre-trained base classes, we propose a novel Dual-Embedding-Space (DES) architecture which decouples the representation learning of base and novel categories into different spaces. Secondly, to mitigate the catastrophic forgetting on the accumulated novel classes, we propose a Sequential Model Fusion (SMF) method, which is able to achieve long-term memory without additional storage cost. Thirdly, to promote inter-task class separation in feature space, we propose a novel regularization technique that extends the classification boundary further away from the previous classes to avoid misclassification. Overall, our framework is simple yet effective and outperforms the previous SOTA with a significant margin of 2.4 points in AP performance. %\hl{SOTA, better?}

\end{abstract}

%%%%%%%%%%%%%%%%%%%%%%%%%%%%%%%%%%%%%%%%%%%%%%%%%%%%%%%%%%%%%%%%%%%%%%%%%%%%%%%%
\section{Introduction}

Object detection~\cite{NIPS2015_FasterRCNN,Bochkovskiy_2020_Yolo4,cascade_rcnn, Li_2021_CVPR} is an fundamental vision task which is required by a lot of robotic applications, such as grasping, manipulation, and navigation~\cite{Karaoguz2019,LiTIE2021,chiu2020probabilistic, Bae2020,Langer2020}, etc. However, different with the standard computer vision task setting that prior large-scale dataset and sufficient computation resource are available for model learning, the successful deployment of object detection on robotic tasks pose many new challenges for the algorithms. First, as robot is desired to adapt and handle new tasks flexibly~\cite{ZhuIROS2020}, the detection task is often involving frequently. As a result, the feature to learn novel classes from only few training samples is highly expected to avoid the time-consuming data labelling, especially for many applications in a low-data regime where the labeled data is rare and hard to collect. Second, robot is also desired to learn additional tasks continuously without forgetting the previous learned knowledge~\cite{ZhuTASE2021}, which demands incremental learning to online reliably register new tasks for a deep model in a continuous manner. Moreover, the model deployment on robotic system is normally with limited computation resource, such as on edge device, etc., which further restricts the incremental learning on hardware limitation.

\begin{figure}
	\centering
	\includegraphics[width=1\columnwidth]{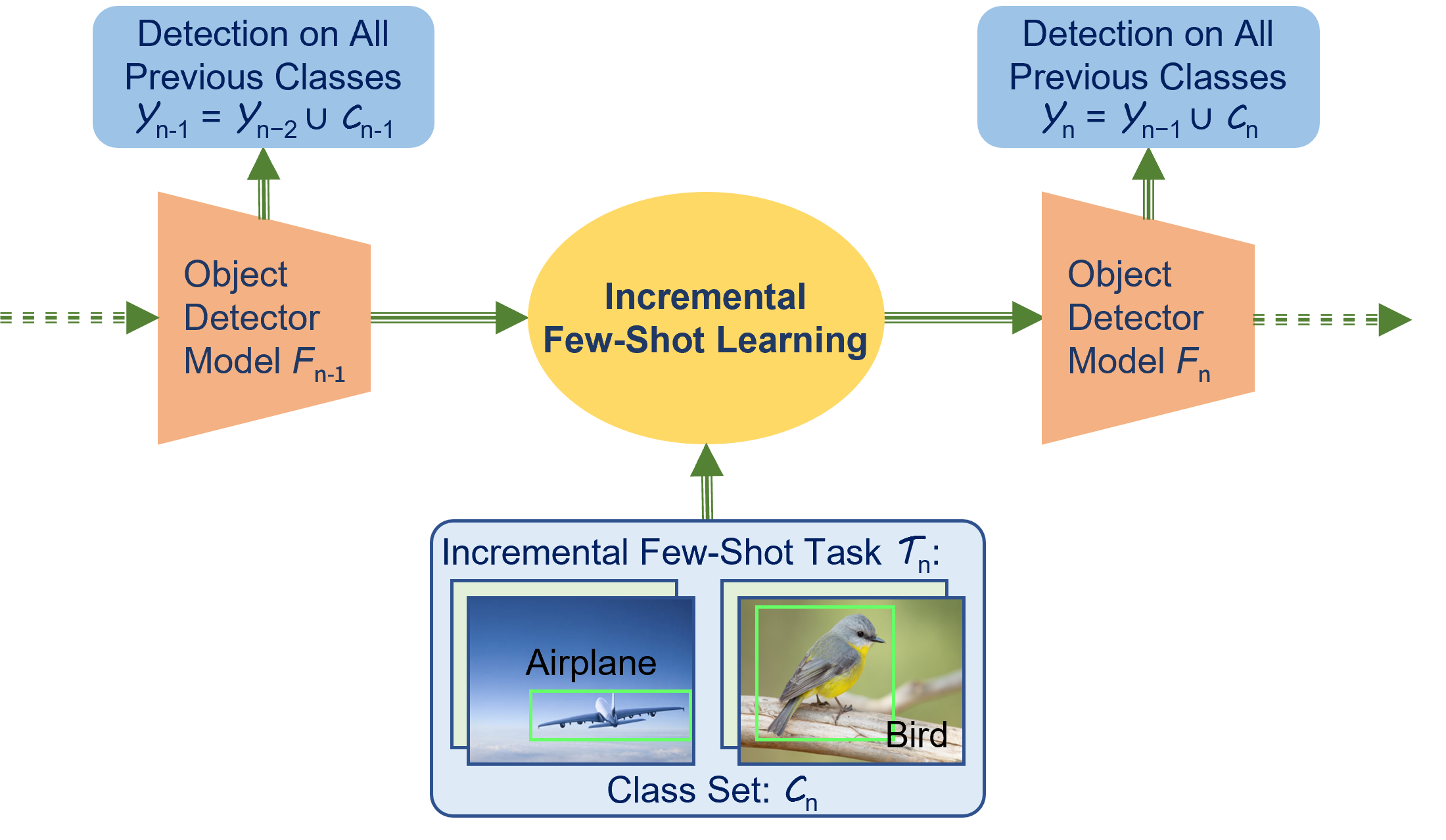}
	\caption{We propose a Class-Incremental Few-Shot Object Detection (CI-FSOD) framework, it can incrementally adapt the online detector model to handle new detection tasks in an expendable manner by using only few-shot samples, while the performance on previous classes are not degraded significantly. Moreover, such model adaptation only utilize few-shot training samples for new classes in current incremental step without accessing the old-class samples, so that the adaptation is very lightweight which is friendly for robotics deployment with computation hardware limitation.}
	\label{fig:ProblemDefinition}
\end{figure}

%based object detection approaches have achieved remarkable results on benchmarks such as Pascal VOC and MS-COCO datasets. This, however, usually requires a large amount of labeled data, which can be labor-intensive and time-consuming to acquire. On the contrary, humans can recognize and locate new objects after observing only one or a few instances. Such ability to learn from few examples is desirable for many applications in a low-data regime where the labeled data is rare and hard to collect. The challenge of detecting novel classes from few samples is usually referred to as Few-Shot Object Detection (FSOD), which is usually framed as either metric learning or meta-learning problems in previous studies. For instance, TFA~\cite{Wang2020} is one of the most representative approaches that adopting metric learning into few-shot detection.  Meta-RCNN~\cite{Yan2019MetaRCNN} propose a meta-learning attention mechanism to emphasize correlated feature channels for better classification. Despite their successes, when extending to novel classes, they require to store and replay a memory buffer for old classes, which often results in heavily model retraining and inevitably hinders their real-world applications.

To address the above practical challenges for robotics society, we propose a novel framework in this work to achieve Class-Incremental Few-Shot Object Detection (CI-FSOD) simultaneously, shown as in Fig.~\ref{fig:ProblemDefinition}. Unlike the conventional FSOD methods, CI-FSOD requires not only to reduce the hardware resource requirement but also to retain the comparable detection performance. Hence, a desirable solution should contain three important aspects. First, the model should learn efficiently only from a continual data stream without catastrophic forgetting. When the model adapts to new tasks, it is not allowed be fine-tuned with previous task data to achieve fast adaption and memory saving. Second, the model should generalize well even when the training samples are very limited for novel classes. Third, previous knowledge gained from the large-scale base training set should be well preserved without dramatic degradation. In view of these, it is noted that many previous approaches only focus on promoting feature representation for novel classes but fail to achieve good knowledge retention on the base classes.

Catastrophic forgetting affects the deep model learning, limiting their capabilities to learn multiple tasks sequentially. To tackle catastrophic forgetting on the large-scale base classes in order to maintain the overall performance, TFA~\cite{Wang2020} only fine-tunes the last two $fc$ layers on a small balanced training set, while the feature extractor is  fixed without adaption. This could be viewed as a practical approach to preserve the previously learned feature distribution for base classes. However, it is unreasonable that the pre-trained feature extractor, which only contains semantic information of base classes, could be used directly to represent unseen novel classes without further fine-tuning. In contrast, when we fine-tune TFA by unfreezing ROI feature extractor, it is surprising to see that the AP performance on novel classes could be significantly improved from $9.9$ to $12.3$ points. Such findings indicate that fine-tuning deep CNN feature extraction layers is crucial for learning unseen concepts since it can obtain better features. However, the worse results on base class (decreasing from $28.8$ to $26.5$ points) reveal that fine-tuning base classes with limited data may inevitably lead to catastrophic forgetting of the original generalizable representations~\cite{BBN}. In this work, to better preserve base classes' performance, we propose to decouple the representation learning of base and novel classes into two independent spaces. In particular, a novel Dual-Embedding-Space (DES) architecture is proposed to take care of both base-class retention and novel-class adaption simultaneously.

CI-FOSD also faces the category confusion on those incrementally encountered novel classes. A common approach Incremental Moment Matching (IMM)~\cite{IMM} proposes to prevent catastrophic forgetting by interpolating weights of different learning tasks, which aims at restricting the parameter for the current task to be nearby the minima of previous tasks. However, IMM cannot achieve long-term memory unless storing model weights for all tasks, which can be highly memory-inefficient to a long sequence of tasks~\cite{longmemory}. On the other hand, in the incremental learning setting, the model adaptation can only access the data of the current task while no data of previous tasks can be stored for fine-tuning. As a result, inter-class separation becomes a challenging problem as it is almost not possible to achieve by only using the few-shot training data of the current-task classes, where the discrimination between classes of different tasks is missing during the training.

In this work, we address the above learning issues from two perspectives, which is integrated with the proposed DES framework. First, networks trained on sequential few-shot tasks are more likely to forget previously gained knowledge. We propose an improved Sequential Model Fusion (SMF) approach for CI-FSOD setting, which exerts a stronger resistance force towards model updating by performing weight merging at each training epoch, so that the training stability can be reinforced and long-term memory can be achieved. Second, a margin-based center loss is incorporated with the proposed CI-FSOD framework to further enhance the feature discriminability as well as the inter-task class separation CI-FSOD. Our approach only stores the representative feature embeddings rather than raw images, so that it keeps the lightweight and memory-saving model adaptation for CI-FSOD, which suitable for robotics deployment on edge device.

%(3)	To better preserve performance on base classes and also incorporate our method with non-incremental settings, we propose to decouple the representation learning of base and novel classes for the CI-FSOD problem. In particular, we propose a novel -Branch Framework to take care of both base knowledge retention and novel knowledge adaption simultaneously.

\section{Related Works}

\begin{figure*}[tb!]
\centering{\includegraphics[width=1.9\columnwidth]{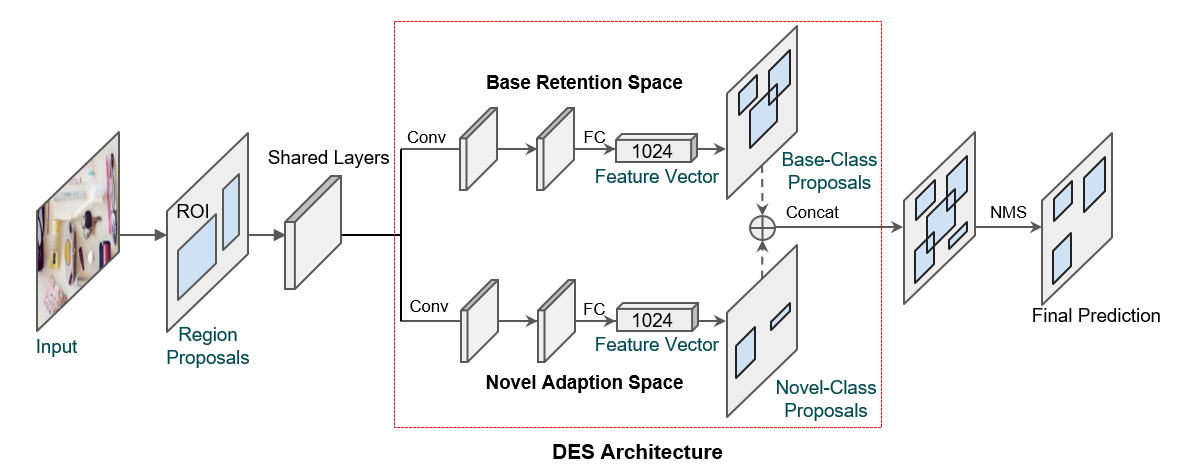}}
\caption{Illustration of the proposed Dual-Embedding-Space (DES) architecture for CI-FSOD. The Base Retention Space only updates its new-added weights on the box classifier in the current task, responsible for preserving the original base-class feature distributions. The Novel Adaption Space jointly updates the last layer of ROI feature extractor and the new-added box classifier weights, which is designed to model the new-coming classes better. Finally, the detected boxes of these two spaces are concatenated together for NMS.}
\label{fig:DBF}
\end{figure*}

\subsection{Incremental Learning}
Recent research that focuses on how to remedy catastrophic forgetting can be roughly grouped from two aspects: data replay and model regularization. The first group of methods usually store a memory of the learned knowledge from historical tasks. For example, iCaRL~\cite{rebuffi2017icarl} builds a class-representative memory with exemplars that are close to the category center. BiC~\cite{bic} propose to learn a bias correction module from an unbiased validation set to eliminate prediction bias between old and new tasks. Regularization-based methods pay attention to stabilize parameters updating during the learning process. For example, EWC~\cite{EWC} exerts a constrain on the alterations of model parameters according to the importance of weights in previous learned tasks given by fisher information. Incremental Moment Matching (IMM)~\cite{IMM} restricts the parameters updating for the current task to be close to the local minima of the previous task, an extra L2 regularization term is added to preserve linear connectivity between multiple tasks. However, to achieve long-term memory mainly, they require storing either Fisher matrix or model weights for each task independently, which is practically infeasible if there are many tasks and the network has millions of parameters.

\subsection{Incremental Few-Shot Learning}
Incremental few-shot learning is attracting increasing attention due to its demands from realistic applications~\cite{Chelsea_2017_MAML,Snell_2017_Prototypical}. However, most of the existing methods are proposed to address the single image classification problem, hence not readily applicable to object detection. Our method falls within the context of the regularization-based learning approach~\cite{IDQ,IMM,EWC}. For example, when learning incrementally, ILDVQ~\cite{IDQ} preserves feature representation of the network on older tasks by using a less-forgetting loss, where response targets are computed using data from the current task. As a result, ILDVQ does not require the storage of older training data. However, this strategy may be inefficient if the data for the new task is too few or just belongs to a distribution different from those of prior tasks.%  In contrast, we address the incremental few-shot learning problem from a new perspective of parameter space. The proposed Stable Moment Matching algorithm strengthens the stability of few-shot adaptation and is more robust to data scarcity.

\subsection{Few-Shot Object Detection}
The most recent few-shot detection approaches are adapted from the few-shot learning paradigm. \cite{Hao2018LSTD} proposes a distillation-based approach with background depression regularization to eliminate the redundant amount of distracting backgrounds. A meta-learning attention generation network is proposed in \cite{Kang_2019_YOLOLS} to emphasize the category-relevant information by reweighting top-layer feature maps with class-specific channel-wise attention vectors. Sharing the same insight, Meta-RCNN~\cite{Yan2019MetaRCNN} applies the generated attention to each region proposal instead of the top-layer feature map. TFA~\cite{Wang2020} replaces the original classification head of Faster-RCNN with a cosine classifier to stabilize the adaptation procedure. %However, existing FSOD methods mainly consider the non-incremental learning setting. When new classes are to be added, they rely on the base training dataset during fine-tuning, which dramatically increases memory space and computation cost, and restricts their scalability into realistic applications such as IoT and robotics.

\section{Methods}

\subsection{Problem Formulation}

The CI-FSOD problem is normally formulated as a two-phase learning task. In the first representation learning phase ($n=0$), a detection model $F_{0}(\cdot|\bm{W_{0}})$ is pretrained on a large set of base classes where $\bm{W_0}$ denotes the learned weight parameters. In the second incremental learning phase ($n>0$), given the sequential new few-shot task $\mathcal{T}_n$, the model updating $F_{n}(\cdot|\bm{W_{n}})$ is performed over multiple stages from $\mathcal{T}_{1}$ to $\mathcal{T}_{n}$. During the $n$-th learning step, the previous category space $\mathcal{Y}_{n-1}$ is expanded with new classes $\mathcal{C}_n$ in $\mathcal{T}_{n}$, so that $\mathcal{Y}_n = \mathcal{Y}_{n-1}\cup \mathcal{C}_n$, where $\mathcal{Y}_{n-1}\cap \mathcal{C}_n=\emptyset$ is assumed for simplicity. The objective of CI-FSOD is to effectively learn a $F_{n}(\cdot|\bm{W_{n}})$ which is capable to detect all the classes in $\mathcal{Y}_{n}$ with high accuracy. In this work, we use ``base'' and ``novel'' classes to differentiate the classes in representation learning phase ($n=0$, not few-shot manner) and incremental learning phase ($n>0$, few-shot manner). Moreover, we use ``new'' and ``old'' classes to denotes the new coming classes in $\mathcal{T}_{n}$ and previous few-shot classes from $\mathcal{T}_{1\,\text{to}\,n-1}$, noted that both of them are belong to novel classes.

\subsection{Overall Framework}

To balance between base-class retention and novel-class adaptation, we propose a Dual-Embedding-Space (DES) architecture that constructs the representations for base and novel classes separately. The schematics of our method is shown as in Fig.~\ref{fig:DBF}, where the base detector here we use Faster-RCNN for illustration example. It is also noted the proposed framework is also compatible to other detectors such as SSD, etc. The speciality of Fig.~\ref{fig:DBF} exists on the two parallel embedding space, termed as the ``Base Retention Space” and ``Novel Adaption Space”. Given a base detector that is pre-trained on a large-scale base set, we duplicate the last convolutional layer, box classifier and box regression head to form the additional embedding space for representing novel classes.

%where our framework is implemented with two parallel embedding space, termed as the ``Base Retention Space” and ``Novel Adaption Space”. 

For Base Retention Space, to fully maintain base-class performance without forgetting, we utilize the pre-trained model parameters to preserve base-class data distribution. In particular, when adapting to a new task, we freeze all category-agnostic parameters of the base detector, while merely updating the fully-connected box classifier to add novel classes. For Novel Adaption Space, to learn semantically rich features for representing novel classes, we fine-tune the feature extraction layers. In the meanwhile, the proposed SMF (Section III.C) is employed to constrain parameter updating and the inter-class regularization is 
used to prevent category confusion. After finishing the training for all tasks, the detection output of these two spaces are concatenated for form the candidate boxes for NMS, as illustrated in Fig.~\ref{fig:DBF}. Such a combination provides a unified prediction of the overall learned classes in CI-FSOD setting.

Since each incremental step is few-shot, the classification weights need to be reliably extended without overfitting. Hence, we adapt weight imprinting~\cite{Hang2018} into object detection to infer new class weights by averaging the ROIs features of all foreground candidates.  Specifically, for the $n$-th learning task, consider its training set $\mathcal{D}_{n}=\{(\mathcal{I}_{i},c_{i})\}$ where $\mathcal{I}_{i}$ denotes the image sample set for new class $c_{i}\in\mathcal{C}_{t}$. For a class $c_{i}$, a fixed number of $k$ foreground ROI features $\bm{\varphi}_{bij}|_{j=1}^{k}$ are sampled from images $I\in\mathcal{I}_{i}$, so that its category-representative proxy $\bm{\varphi}_{bi}$ for class $c_{i}$ can be approximated by $\bm{\varphi}_{bi}=\frac{1}{k}\sum_{j=1}^{k}\bm{\varphi}_{bij}$. To extend the classifier weight matrix, $\bm{\varphi}_{bi}$ will be added into the current weight collection $\mathcal{W}_{n}=\mathcal{W}_{n-1}\cap\{\varphi_{bi}\,|\,c_{i}\in\mathcal{C}^{n}\}$ in order to provide a better initialization point than random initialization  for the encounted classes in task $n$

\subsection{Sequential Model Fusion (SMF)}
%Catastrophic forgetting is one of the main challenges for CI-FSOD. During the long-sequential incremental learning, data for previous tasks are often unavailable due to the hardware resource limitation or privacy issues~\cite{incrementalOBJECTDETETION}, which causes the performance of the model on previous tasks to degrade dramatically. Regularization-based approaches such as EWC~\cite{EWC} and mean-IMM~\cite{IMM} preserve previous knowledge by preventing the parameter updating for the current task from going far away from the previous optimum. In this work, an improved approach, i.e., Stable Moment Matching, is proposed for CI-FSOD with simplicity and effectiveness, which is compatible and integrated into the dynamic learning branch of the proposed DBF framework.

\begin{figure}
	\centering
	\includegraphics[width=1\columnwidth]{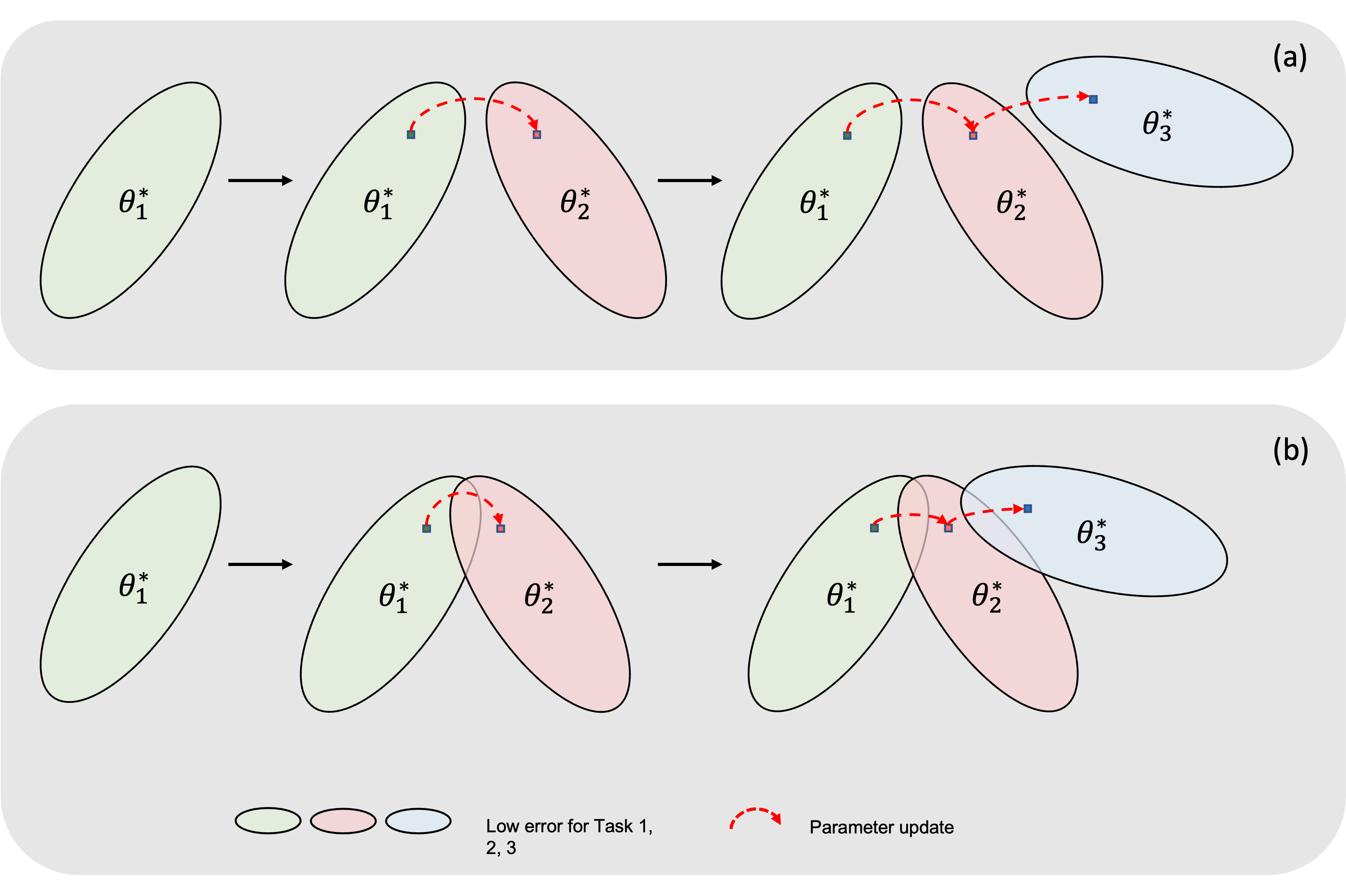}
	\caption{(a) In data-rare cases, the low-error surface around each local minima is sharp and narrow. Thus, the likelihood for these low-error surface overlapping becomes very small when the learning continues, dramatically increasing the difficulty of finding optimal solutions that perform well on different tasks. (b) We restrict the magnitude of parameter displacement for each task,so the obtained local minima can stay closer with each other.}
	\label{fig:WeightDisplacement}
\end{figure}

When two sequential tasks start from the same initialization, IMM assumes that one possible optimum performs well for both tasks should lie on the interpolation line of the individual two solutions. However, long-term memory cannot be achieved unless preserving all previous model weights. Thus, the original IMM only shows good result at the first a few incremental steps, but performance rapidly decays in the following steps. Regarding this, our intuition is that, the distance between any two local minima should be small enough, so that the nearby low-error region of each task’s local minima can overlap with each other when the learning continues. In this work, an improved approach, i.e., Sequential Model Fusion (SMF), is proposed for CI-FSOD with simplicity and effectiveness, which is compatible and integrated into the proposed DES framework.
%\begin{figure}
%	\centering
%	\includegraphics[width=1\columnwidth]{SMF.png}
%	\caption{Illustration of the proposed Stable Moment Matching (SMF). Knowledge base is obtained by only fine-tuning the new-added weights in the box classifier. SMF simply interpolates the parameters of two models during each iteration. Once finishing the current task, we use the last interpolated model as the new knowledge base for next task. }
	%\setlength{\belowcaptionskip}{-2pt}
%	\label{fig:SMF}
%\end{figure}

Assume the local minima obtained through SGD from two continual tasks $\mathcal{T}_{n-1}$ and $\mathcal{T}_{n}$ are $\bm{w}^{n-1}$ and $\bm{w}^{n}$, respectively, where $\bm{w}$ represents model parameters contained in ROI feature extractor, box classifier, and box regressor in CI-FSOD. IMM simply adopts parameter interpolation as,
\begin{equation}
\label{eq:IMMinterpolation}
\begin{aligned}
\bm{w}^ \ast = \alpha\bm{w}^{n-1}+(1-\alpha)\bm{w}^{n},
\end{aligned}
\end{equation}
\normalsize
where $\alpha$ is the interpolation ratio. If $\alpha=0.5$, $\bm{w}^ \ast$ becomes the mean of $\bm{w}^{n-1}$ and $\bm{w}^{n}$. As we can see, IMM has a moderate effect of weight-consolidation as it can restrict the magnitude of the overall trajectory to be one half, however, that’s still not enough for preserving long-term memory through muti-step incremental tasks. Thus, our aim is to seek stronger consolidation effect, which can be simply achieved by performing IMM more times, i.e, at the end each training epoch, rather than only at the end of training. As a result, the obtained local minima can be closer and overlap with each other.
\normalsize
\begin{figure}[tb!]
	\centering
	\includegraphics[width=0.9\columnwidth]{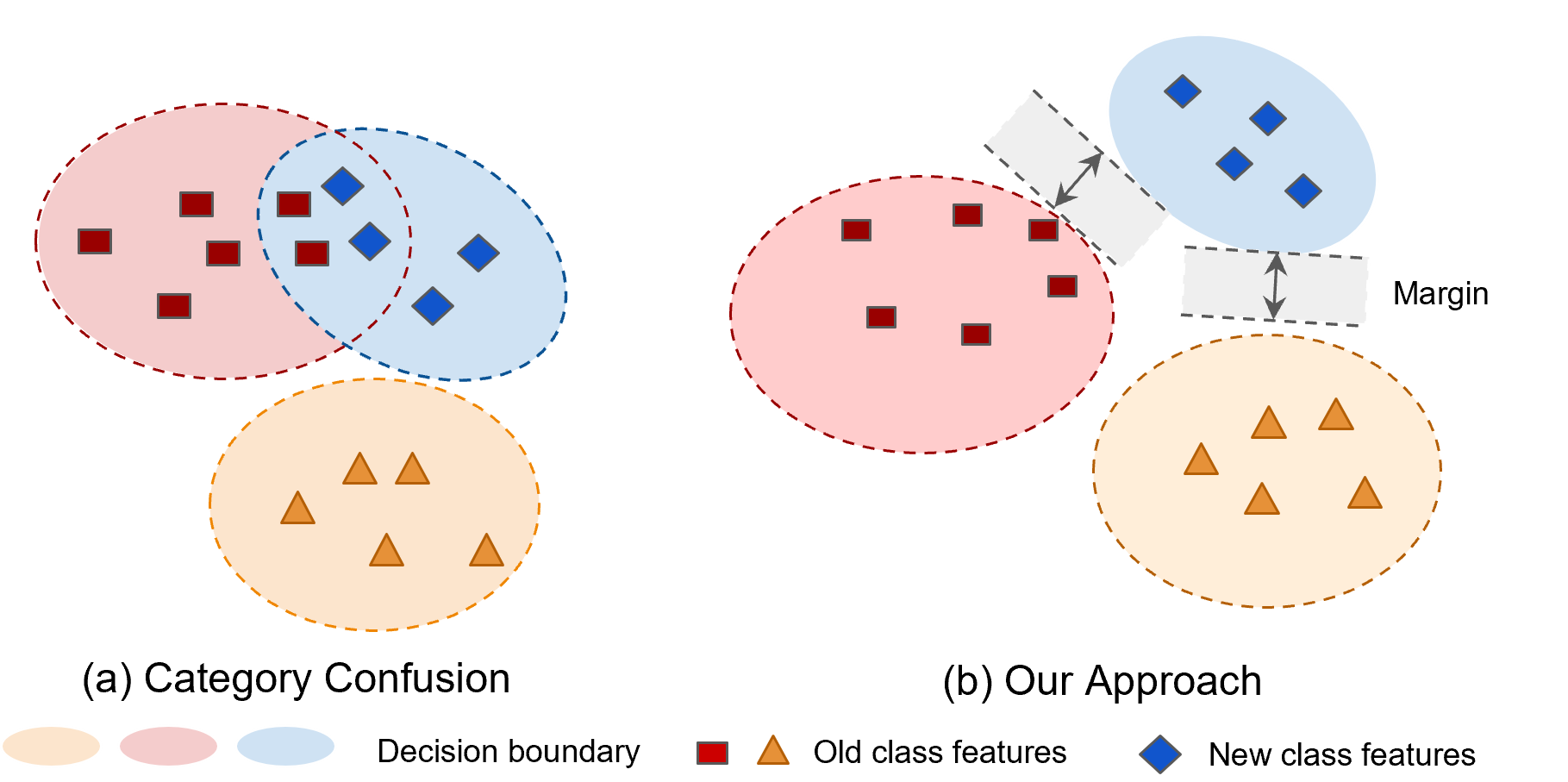}
	\caption{(a) Category confusion. (b) The proposed inter-task class discrimination method promotes a more compact representation with margin for all encountered classes.}
	\label{fig:featuredrift}
\end{figure}

Specifically, the representation learning of each task are decoupled into two phases: classifier learning phase $Phase_{fc}$ and representation learning phase $Phase_{ex}$. In $Phase_{fc}$, assuming the model weights obtained from the previous task $\mathcal{T}_{n-1}$ is denoted as $\bm{w}^{n-1}$, when a new task $\mathcal{T}_{n}$ comes, we adopt the commonly used weight imprinting strategy ~\cite{Hang2018} to initialize the classification weights for the new classes contained in $\mathcal{T}_{n}$. Then, we freeze the backbone, RPN and ROI feature extractor, and only fine-tune the new imprinted weights in the $fc$ classifier as well as the box regressor. The obtained model $\bm{w}^{n}_{fc}$ is then used as the first initialization point for the next representation learning phase $Phase_{ex}$, i.e., $^{init}\bm{w}^{n}_{0}\leftarrow \bm{w}^{n}_{fc}$, where $^{init}\bm{w}^{n}_{t}$ denotes the initialization weight for $t$-th training epoch. In $Phase_{ex}$, gradient descent optimization starts from $^{init}\bm{w}^{n}_{0}$, and after finishing the $t$-th training epoch, the obtained model is denoted as $\bm{w}^{n}_{t}$. The linear interpolation is conducted for every epoch between its current initialization point $^{init}\bm{w}^{n}_{t}$ and the resulted after-epoch-training weight $\bm{w}^{n}_{t}$, the merged linear interpolation weights $^{intpl}\bm{w}^{n}_{t}$ for next epoch is calculated as,
\begin{equation}
\label{eq:interpolation}
\begin{aligned}
^{intpl}\bm{w}^{n}_{t} = ^{init}\bm{w}^{n}_{t}\cdot \alpha +\bm{w}^{n}_{t}\cdot(1 -\alpha),\quad t\geq0
\end{aligned}
\end{equation}
\normalsize
where $\alpha$ is the interpolation ratio. Iteratively, the obtained interpolated weights $^{intpl}\bm{w}^{n}_{t}$ is then used as the initialization point for the next epoch, where $ ^{init}\bm{w}^{n}_{t+1}=^{intpl}\bm{w}^{n}_{t}$. The proposed general updating rule can be summarized as the following,
\begin{equation}
\label{eq:interpolationall}
\begin{aligned}
\textbf{for} \ &t\text{-th epoch ($t\geq1$):}\\
&a,\text{ set}\ ^{init}\bm{w}^{n}_{t} = {^{intpl}\bm{w}^{n}_{t-1}}\ \text{as training start}\\
&b,\text{ train\ one\ epoch\ using\ SGD:}\ \bm{w}^{n}_{t} \stackrel{SGD}{\longleftarrow} {^{init}\!\bm{w}^{n}_{t}}\\
&c,\text{ merge}\ ^{intpl}\bm{w}^{n}_{t} = {^{init}\bm{w}^{n}_{t}}\cdot \alpha +\bm{w}^{n}_{t}\cdot(1 -\alpha)
\end{aligned}
\end{equation}

%With a constant learning rate, the interpolation ratio $\alpha$ mainly determines the displacement of the total trajectory. A smaller $\alpha$ may increase  the overall displacement from the beginning of training to bring more plasticity. However, using a small interpolation rate means applying a larger update to the model weights for a sequential learning problem. Therefore, the resultant model may learn fast but also forget quickly. Hence, we propose an adaptive interpolation ratio that starts with a small interpolation rate at the beginning of training. Then, we slightly increase the interpolation rate for each subsequent task to stabilize the optimization trajectory and prevent forgetting. Precisely, the adaptive interpolation ratio $\alpha$ for the $n$-th task in $q$-th training epoch is calculated as,
%\begin{equation}
%\begin{aligned}
%\alpha(n, q) = (r_{step}\cdot n +r_{base})\cdot q \Big{/} N_{epoch},
%\end{aligned}
%\end{equation}
%where $r_{base}$ and $r_{step}$ are the base rate and step increasing rate. $N_{epoch}$ represents the total number of training epochs for each task. Note that the current epoch number $q$ will be reset to $0$ at the beginning of each task.

\subsection{Inter-Task Separation Loss}

Eliminating the inter-task class ambiguities is challenging as it requires storing data for all old classes in previous tasks, which may cause data privacy or hardware memory issues for CI-FSOD. In this work, we propose a more feasible approach that only stores representative feature embeddings rather than raw images, which is more memory-efficient and privacy-secure. Although feature space generally suffers from drifting along each incremental step, such drift can be effectively restricted to a minimal amount by our proposed SMF regularization. Therefore, the last layer feature vectors can be used as a practical approximation to represent old-class feature distributions.

Our approach is based on the observation that old classes' embeddings are easy to confuse with new classes in future tasks. Therefore, a margin-based separation loss can be used to expand decision regions for old classes and encourage a more compact representation for new classes. As a result, the model's generalization capability can be improved. Specifically, the foreground ROI feature embeddings are sampled during each historical incremental task ($\mathcal{T}_{i=1\,\text{to}\,n-1}$) and stored as $\bm{X}_{old}=\{f(\bm{x})\}$, where $f(\bm{x})$ denotes the extracted feature for the corresponding ROI proposals. Given a new-task few-shot dataset $\mathcal{T}_n$, we randomly sample a fixed number of their foreground ROI features $\bm{X}_{new}=\{f(\bm{x})\}$ through each image. In the meantime, an equal number of historical feature embeddings are sampled from $\bm{X}_{old}$ for each old class. With the cosine similarity, if a ROI feature $x$ belongs to the new classes, we use the normalized weights of its ground-truth class weight vector $\theta_{new}^{+}$ as a positive template, and the negative template $\theta_{old}^{-}$ is selected from the old classes' weight vector that yields the highest similarity to $f(\bm{x})$, while the same arrangement is applied to those old-class feature embeddings. Mathematically, the proposed inter-task separation loss can be formulated as
\begin{equation}
\begin{aligned}
 L_{mgn}=& \sum_{\bm{x}_n \in {New}} \max \big(m- \kappa \theta_{new}^{+} f(\bm{x}_n)+ \kappa \theta_{old}^{-} f(\bm{x}_n),\ 0 \big)\\
 &+\sum_{\bm{x}_{o} \in {Old}} \max \big(m- \kappa \theta_{old}^{+} f(\bm{x}_{o})+ \kappa \theta_{new}^{-} f(\bm{x}_{o}),\ 0 \big) \end{aligned}
\end{equation}
\normalsize
where $m$ is a margin, $\kappa$ is a  scale parameter used to ensure the convergence of training~\cite{cosface}, and $\theta_{new}^{+}$ and $\theta_{old}^{+}$ are the positive templates for new-class and old-class proposals, respectively, using the weight vectors for the ground-truth classes. $\theta_{old}^{-}$ and $\theta_{new}^{-}$ are the negative templates for new-class and old-class proposals, respectively, using the highest-similarity weight vectors from the other groups, i.e., old and new classes.

% \textbf{Inference}
% Upon finishing of training, FSCN can be used to refine all proposals from the last stage of Faster-RCNN. Specially, we propose to fuse their classification score via:
% \begin{equation}
% \begin{aligned}
% s=s_{detector} \odot s_{corrector}
% \end{aligned}
% \end{equation}
% \normalsize
% where $s_{detector}$ and $s_{corrector}$ denote the softmax classification score of the base detector and FSCN, respectively.

\section{Experiments}

\subsection{Implementation Details}
We use Faster-RCNN as our base detector and an ImageNet pretrained Resnet-50~\cite{ResNet} is employed as the backbone. For the pre-training phase, a detection model is optimized with Stochastic Gradient Descent for the first five epoch, with a minibatch size of 16, learning rate of 0.02, momentum of 0.9, and weight decay of 0.0001. For the incremental fine-tuning phase, we sequentially  add one class per task. Traing epoch for each task is set to be 7. A smaller learning rate of 0.005 is adopted for the two spaces. For the adaptive interpolation ratio, we set $\alpha(n, t)=0.5$. For $L_{mgn}$, we use $\kappa=10$ to ensure the training convergence. For the loss importance $L_{total}= L_{ce}+\lambda_{1} \cdot L_{mgn}+  \lambda_{2} \cdot L_{reg}$, we use $\lambda_{1} = 1$ and $\lambda_{2}=0.001$. Our implementation is based on Pytorch with 4 GTX 2080TI.

\subsection{Results on MS-COCO Dataset}
%\subsubsection{Experimental Setup}
%To evaluate our approach, we follow the common practice of $k$-shot but with the key difference that no image of the previous task is stored. In particular, our approach is evaluated on the commonly used object detection benchmarks Pascal VOC~\cite{PascalVOC} and MS-COCO~\cite{COCO}, following the same experiment setting with Meta-RCNN~\cite{Yan2019MetaRCNN}.

%For the first same-domain evaluation, we train our model on the union of the 80K train set and 35K trainval set of MS-COCO, and test on the 5K minival set. For the total 80 categories, 20 categories that are overlapped with Pascal VOC are selected as novel classes, and the remaining 60 classes serve as base classes. It is worth to note that base-set images may also contain instances from novel classes. However, we do not provide any annotations for novel-class instances during pretraining. Upon the arrival of each novel class, the model can only access to the annotation of $k$ instances, where $k=\{1,5,10,30\}$ for all experiments. For the cross-domain evaluation from MS-COCO to VOC, the model is first trained on MS-COCO, and then evaluated on VOC2007 test set. Different from the first experiment that focuses on evaluating cross-category model generalization, this setup is further to appraise the cross-domain generalization ability.
% generalization ability.

To evaluate our approach, we follow the common practice of $k$-shot but with the critical difference that no image of the previous task is stored. In particular, we split the 80 class in MS-COCO into 60 base class and 20 novel class. During incremental learning, the 20 novel class comes in a sequential manner, and each incremental step only contains one particular class, the overall training is finished after 20 incremental steps. As shown in Table~\ref{table:incremental COCO}, we have compared our approach to multiple baselines. The first two baselines, ``Feature-Reweight" ~\cite{Kang_2019_YOLOLS} and ``ONCE" ~\cite{Once} are meta-learning approaches. The third baseline denoted as ``Imprinting" denotes the original imprinting approach used for image classification~\cite{Hang2018}. For the fourth baseline, ``TFA-fc," we extend the original TFA, which was developed for a non-incremental setting into a sequential learning manner. In particular, upon each task arrival, we only fine-tune the new-imprinted weights of each novel class for that task. We further unfreeze the ROI feature extractor and box regressor for fine-tuning in the following baseline denotes as ``TFA-ex." In addition, we also include the commonly-used method Incremental Moment Matching into our baseline methods, where denotes as ``TFA-ex + IMM". To evaluate our proposed DES, we only fine-tune the new-imprinted $fc$ weights for the current task for the base retention space, while all other structures are fixed. For the novel adaption space, we fine-tune the ROI feature extractor, box regressor, and the new-imprinted weights (only the current task) for the box classifier with multiple strategies.

Regarding the results, we have several observations. (1) The accuracy of the existing meta-learning approaches is still far away from satisfaction. Although fine-tuning meta learners can further improve them, the episodic learning scheme is memory inefficient when the number of classes increases, making them not feasible for real-world CI-FSOD scenarios. (2) Compared with the baseline TFA-fc, unfreezing ROI feature extractor (TFA-ex) leads to worse results, which indicate that naively fine-tuning the feature representation with sequential few-shot data may cause aggravated catastrophic forgetting. (3) IMM can relieve such aggravated catastrophic forgetting to some extent. However, compared with the strong baseline TFA-fc, its effectiveness is still marginal. (4) The proposed SMF approach outperforms the original IMM approach by 1.1 points under 10-shots, which indicates that restricting the overall parameter displacement is crucial for preventing catastrophic forgetting. (5) The proposed inter-task class separation regularization consistently performs better than training with traditional cross-entropy loss, indicating the importance of inter-task class discrimination. In the end, our proposed methods gain +1.8 AP for 5-shot and +2.4 AP for 10-shot above current SOTA, which is more significant than the gaps between any previous advancements.

\subsection{Non-Incremental Results on Pascal VOC} We provide the AP50 performance on VOC 07 test set in Table~\ref{table: VOC}. All methods are fine-tuning on a balanced training set which contains equal number of samples for both base and novel classes. As we can see, our methods consistently outperform meta-learning approaches (Feature-Reweight,Meta-RCNN) by around 10--20 points in fewer-shot setting, and achieve better results when comparing with the baseline approach TFA-fc~\cite{Wang2020}. As we can see, although further unfreezing the ROI feature extractor (TFA-ex) can improve novel-class performance due to better features, fine-tuning on a small balancing training set may aggravate forgetting on base classes (comparing with TFA-fc). With the proposed SMF approach, such forgetting can be effectively solved. Furthermore, performance can be effectively improved by our proposed Inter-task class separation loss, which is benefited from the more discriminative classification boundaries.

\subsection{Results on MS-COCO to Pascal VOC} 
The cross-dataset setting is evaluated for the proposed method from MS-COCO to VOC, the model is first trained on MS-COCO as the base set pre-train, and then few-shot performance is evaluated on VOC2007 test set. This setup is further to appraise the cross-domain generalization ability. The results in Table~\ref{table:COCO to VOC} confirm the generalization advantages of our method when transfers to a test domain different from the training one.

% \begin{table}
%     \caption{Comparison with CI-FSOD baselines on MS-COCO.}
%     \label{table:incremental COCO}
% \centering
% \setlength{\tabcolsep}{1.5mm}{

%  \begin{tabular}{*{22}{l}}

%     \toprule
%         & \multirow{1}{*}{{Shots}}  & \multirow{1}{*}{{Methods}}

%     &        & Novel AP    & Base AP    \\
%     \midrule

%   &  & Feature-Reweight &   &0.8    & 3.3      \\
%     & 5 & ONCE   &   &1.0    &17.9&        \\
%     %  &  & Original IMP  &   & 5.4    &    28.0    \\
%   &  & Imprinting &   & 4.3    &28.6        \\
%   &  & TFA-fc & & 4.7 &  28.8        \\
%   &  & TFA-ex & & 3.9 &  20.3        \\
%   &  & TFA-ex + mean-IMM & & 5.2 &  23.8        \\
%   &  & DBF (ours) & & 3.9 &  28.7        \\
%   &  & DBF + mean-IMM (baseline) & & 5.1 &  28.8        \\
%   &  & DBF + SMF (ours) & & 6.2 &  28.8        \\
%     &  & DBF + SMF + Inter-Sep (ours) &       &  \textbf{7.1}  &\textbf{28.8}    \\
%       \midrule

%   &  & Feature-Reweight   &   &1.5    &3.7&        \\
%     & 10 & ONCE    &   & 1.2   &17.9&        \\
%  %&  & Original IMP &&6.4   & 27.9    &        \\
%   &  & Imprinting &   &5.8     &  28.5  &    \\
%   &  & TFA-fc & & 6.2 &  28.7        \\
%   &  & TFA-ex & & 5.1 &  19.5        \\
%   &  & TFA-ex + mean-IMM & &6.5  &  22.6        \\
%   &  & DBF (ours) & & 5.0 &  28.7        \\
%   &  & DBF + mean-IMM (baseline) & & 6.5 &  28.7        \\
%   &  & DBF + SMF (ours) & & 7.3 &  28.7        \\
%     &  & DBF + SMF + Inter-Sep (ours) &  & \textbf{8.7}   &  \textbf{28.7}&        \\
%     \bottomrule
%     \end{tabular}}
% \end{table}

\begin{table}
    \caption{Comparison with CI-FSOD baselines on MS-COCO.}
    \label{table:incremental COCO}
\centering
\setlength{\tabcolsep}{1.5mm}{
\begin{tabular}{clcccc}
\toprule
\toprule
\multicolumn{2}{l}{\multirow{2}{*}{}}                   & \multicolumn{2}{c}{5-Shot}   & \multicolumn{2}{c}{10-Shot}   \\
\cline{3-6}
\multicolumn{2}{c}{Method}                                    & \makecell[c]{Novel\\ AP}     & \makecell[c]{Base\\ AP}       & \makecell[c]{Novel\\ AP}     & \makecell[c]{Base\\ AP}        \\
\hline
\multirow{10}{*}{} & Feature-Reweight             & 0.8          & 3.3           & 1.5          & 3.7            \\
                         & ONCE                         & 1.0          & 17.9          & 1.2          & 17.9           \\
                         & Imprinting                   & 3.8          & 28.6          & 5.8          & 28.5           \\
                         & TFA-fc                       & 4.5          & 28.8          & 6.2          & 28.7           \\
                         & TFA-ex                       & 3.4          & 20.3          & 5.1          & 19.5           \\
                         & TFA-ex + IMM            & 4.6          & 23.1          & 6.5          & 22.6           \\
                         & DES (ours)                   & 3.2          & 28.7          & 5.0          & 28.7           \\
                         & DES + IMM (baseline)    & 4.7          & 28.8          & 6.5          & 28.7           \\
                         & DES + SMF (ours)             & 5.6          & 28.8          & 7.3          & 28.7           \\
                         & DES + SMF + Inter-Sep (ours) & \textbf{6.3} & \textbf{28.8} & \textbf{8.6} & \textbf{28.7} \\
                         \bottomrule
\end{tabular}}
\end{table}

\begin{table}[ht]
\small
\caption{Non-Incremental learning results on Pascal VOC}
\centering
\setlength{\tabcolsep}{1.5mm}{

 \begin{tabular}{*{22}{l}}

    \toprule
    \toprule
        & \multirow{1}{*}{{Shots}}  & \multirow{1}{*}{{Methods}}

    &        & \makecell[c]{Novel\\ AP50}    & \makecell[c]{Base\\ AP50}    \\
    \midrule

  &  & Feature-Reweight  &   & 33.9    & -       \\
   &  & Meta-RCNN &   &45.7   & -      \\
    & 5 & TFA-fc   &   &53.1   &\textbf{67.9}  &        \\
    &  & TFA-ex &  &  \textbf{55.8}  &64.3    \\
     &  & TFA-ex + SMF +Inter-sep (ours) &  &  55.7  &67.5    \\
      \midrule
      &  & Feature-Reweight &  & 47.2   &     -   \\
   &  & Meta-RCNN   &   &51.5   &-&        \\
    & 10 & TFA-fc    &   & 58.7  &\textbf{68.1}&        \\
    &  & TFA-ex &  & 61.1   &  65.3&        \\
    &  & TFA-ex + SMF +Inter-sep (ours) &  & \textbf{61.2}   &  67.9&        \\
    \bottomrule
    \end{tabular}}

\label{table: VOC}
\end{table}

\begin{table}
    \caption{Comparison with baselines from MS-COCO to Pascal VOC.}
\centering
\setlength{\tabcolsep}{1mm}{

 \begin{tabular}{*{22}{c}}
    \toprule
    \toprule
        & \multirow{1}{*}{{Methods}}

    &        & 5-Shot & 10-Shot         \\
    \midrule

  %& MAML  &   &   0.6  &   1.0  \\
  & ONCE   &   & 2.4 & 2.6     \\
  & Imprinting   &   & 3.9 & 10.6     \\
  & TFA-fc   &   & 4.3 & 12.7     \\
   & DES + SMF + Inter-Sep (ours) &   &  \textbf{6.5}  & \textbf{15.1}       \\
    \bottomrule
    \end{tabular}}
\label{table:COCO to VOC}
\end{table}

\section{Conclusion}
In this work, we propose a generic learning scheme for CI-FSOD, which incrementally adapt the online detector model to handle new detection tasks in an expendable manner by using only few-shot samples. Our approach also addresses the performance degradation issue of CI-FSOD on previous classes by the proposed DES and SMF methods. Moreover, an Inter-Task Class Separation loss is proposed to promote a better separation margin between old and new classes without accessing the previous old-class training images, so it is very suitable for robotics deployment on edge device. The effectiveness is validated by extensive experiments, where our approach yields the state-of-the-art performance and outperform the previous algorithm with a significant margin.

\bibliographystyle{IEEEtran}
\bibliography{ICRA_final}%Reference_xbib

\end{document}